\documentclass[10pt,twocolumn,letterpaper]{article}

\usepackage{multirow}
\usepackage[table,dvipsnames]{xcolor}
\usepackage{microtype}
\newcommand{\faintrule}{%
  \arrayrulecolor{black!20}\midrule\arrayrulecolor{black}%
}

\makeatletter
\def\@fnsymbol#1{\ensuremath{\ifcase#1\or \ast\or \dagger\or \ddagger\or
   \mathsection\or \mathparagraph\or \|\or **\or \dagger\dagger
   \or \ddagger\ddagger \else\@ctrerr\fi}}
\makeatother

\definecolor{mygreen}{HTML}{E2F0D9}
\usepackage[pagenumbers]{fisherdistill}

\definecolor{linkblue}{rgb}{0.21,0.49,0.74}
\usepackage[pagebackref,breaklinks,colorlinks,allcolors=linkblue]{hyperref}
\hypersetup{pdftitle={FisherDistill: Diffusion Distribution Distillation via Closed-Form Fisher Critics},
            pdfauthor={Mahesh Bhosale, Yuanhao Zhai, Siwei Lyu, Junsong Yuan, David Doermann, Defang Chen}}

\title{FisherDistill: Diffusion Distribution Distillation via Closed-Form Fisher Critics}

\author{
Mahesh Bhosale$^{1}$ \quad
Yuanhao Zhai$^{2}$\thanks{This work was conducted in personal capacity and Meta was not involved in the work.} \quad
Siwei Lyu$^{1}$ \quad
Junsong Yuan$^{1}$ \\
David Doermann$^{1}$ \quad
Defang Chen$^{3}\thanks{Corresponding author}$\\[3pt]
$^{1}$University at Buffalo, SUNY \qquad
$^{2}$Meta \qquad
$^{3}$University of California, Berkeley\\
}

\begin{document}
\maketitle
\begin{abstract}
Recent advances in diffusion distillation have enabled high-quality image generation in only one or a few network function evaluations (NFEs). Despite their efficiency, aggressively distilled generators often retain a residual distribution gap from their multi-step teachers, motivating recent methods to introduce adversarial refinement with a learned discriminator. In this work, we ask whether such refinement necessarily requires a learned adversarial discriminator. We introduce \textbf{FisherDistill}, a simple alternative based on Fisher linear discriminant analysis. Using intermediate representations from the frozen teacher, we estimate the first- and second-order statistics of real and generated samples and derive a closed-form Fisher direction that maximally separates their feature distributions relative to within-class variation. This direction provides a lightweight refinement signal that complements the original distillation objective without requiring an additional adversarial discriminator or its optimization loop. Across multiple distillation frameworks, our approach consistently improves generation quality over their respective distilled baselines and substantially closes the gap to adversarially refined counterparts. 
Code: \url{https://bhosalems.github.io/FisherDistill/}
\end{abstract}
    
\section{Introduction}
\label{Introduction}

Diffusion-based generative models~\cite{ddpm, ddim, ramesh2022hierarchical, lipman2022flow, ldm, dit, sd3} have emerged as a powerful paradigm for high-fidelity content generation, achieving remarkable performance across images~\cite{ldm, dit, sd3}, videos~\cite{vid_diffusion, ho2022imagen}, audio~\cite{liu2023audioldm}, and scientific domains~\cite{watson2023de, bhosale2025pathdiff}. Recent architectural, scaling, and formulation advances have further improved sample quality and realism~\cite{sd3, dit}. However, high-quality generation still requires iterative sampling with repeated network evaluations, incurring a high inference cost~\cite{ddpm, ddim}. Training-free methods reduced this cost through shorter deterministic trajectories~\cite{dpm,chen2024trajectory}, while consistency models enabled direct few-step generation by learning mappings across the diffusion trajectory~\cite{song2023consistency,lcm}. Building on these advances, diffusion distillation compresses the sampling trajectory of a pretrained diffusion model into a one- or few-step generator~\cite{dmd,dmd2,SiD,1_x_distill}. In particular, recent distribution-matching~\cite{dmd,dmd2} and score-identity distillation~\cite{SiD,SiDA} approaches have demonstrated that such aggressively distilled generators can retain much of the quality and diversity of their multi-step teachers, making few-step generation an increasingly practical alternative to iterative sampling.

\begin{figure}[t]
    \centering 
    \includegraphics[width=\columnwidth]{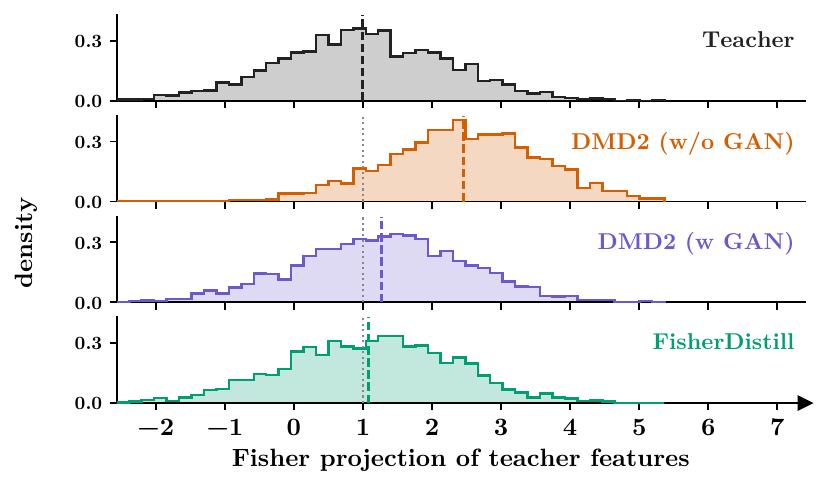}
    \caption{\textbf{Distribution gap in frozen teacher features on ImageNet-64.} One-step DMD2 w/o GAN drifts away from the multi-step teacher, while our closed-form Fisher refinement brings it back to the teacher. Features are held-out Fisher projections of EDM-teacher. Dashed lines mark means.}
    \label{fig:frozen_img_features}
\end{figure}

\begin{figure*}[t]
    \centering
    \includegraphics[width=\textwidth]{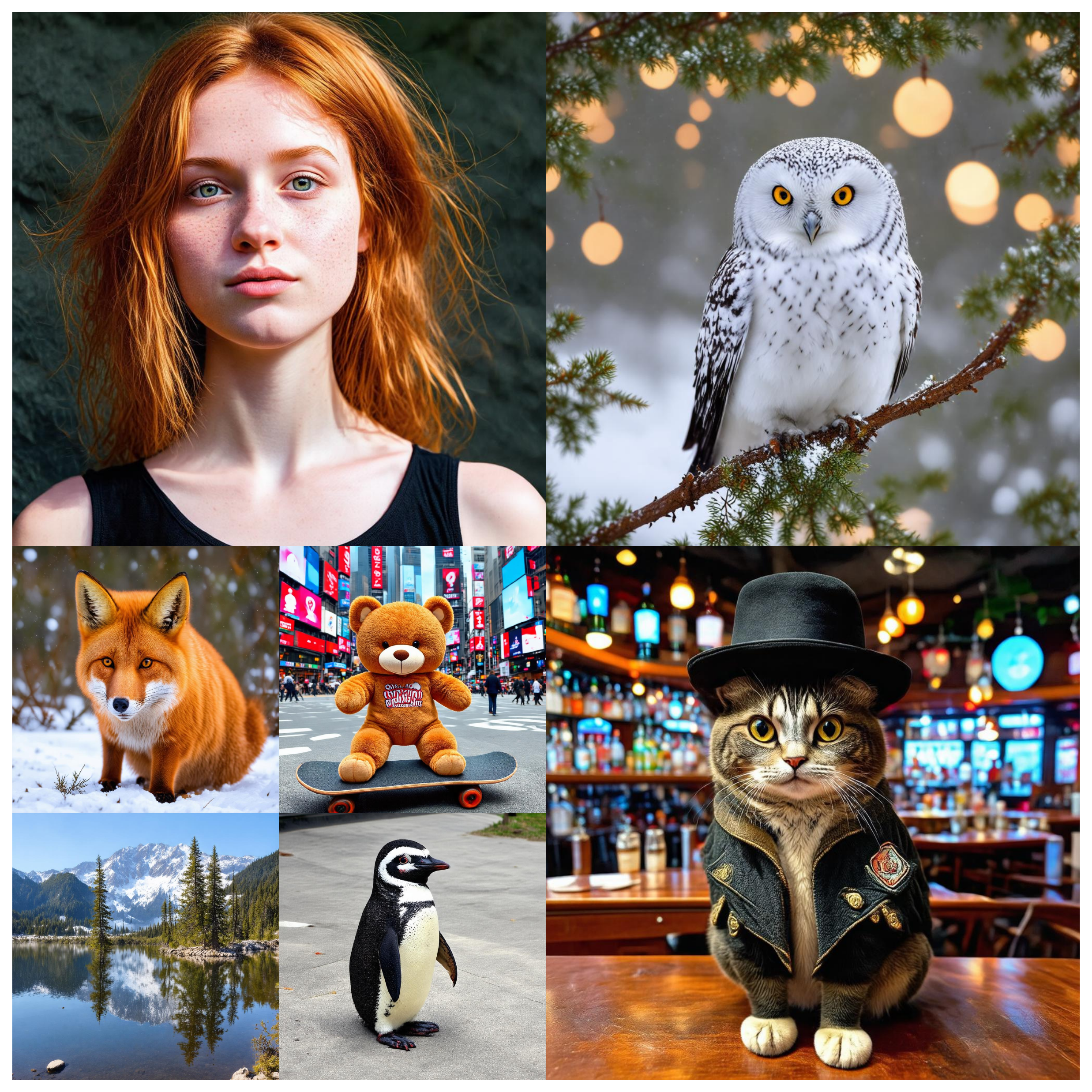}
    \caption{\textbf{1024$\times$1024 samples generated by our 2-step FisherDistill} built on 1.x-Distill Stage-I~\cite{1_x_distill} and distilled from SD3-Medium~\cite{sd3}.}
    \label{fig:Qual_fig1}
\end{figure*}

Despite this progress, aggressive one- or few-step distillation can leave a residual distribution gap from the multi-step teacher (\cref{fig:frozen_img_features}). Recent work shows that adding adversarial supervision to an already distilled generator improves sample quality~\cite{dmd2,SiDA,1_x_distill}. This motivates a coarse-to-fine view of distillation: a base distillation objective first learns most of the teacher distribution, while adversarial supervision refines the remaining discrepancies. When used for such refinement, however, adversarial supervision requires a jointly learned discriminator and an additional min-max objective, inheriting the complexity and limited scalability of adversarial training~\cite{twinflow, dpdmd}.

This limitation raises a natural question of whether closing the residual distribution gap requires a learned adversarial discriminator. We show that it does not (\cref{fig:frozen_img_features,fig:Qual_fig1}). Instead, we exploit the frozen teacher's representation space and estimate the first- and second-order statistics of real and generated features. From these statistics, we derive a closed-form Fisher critic that identifies the direction of maximal separation between the two feature distributions relative to their within-class variation. This direction provides a lightweight refinement signal that complements the original distillation objective, requiring neither a learned discriminator nor adversarial optimization. Our approach thus retains the coarse-to-fine principle while replacing learned adversarial refinement with a closed-form Fisher critic.

\noindent Our main contributions are summarized as follows:
\begin{enumerate}
    \item We introduce \textbf{FisherDistill}, a non-adversarial refinement method that replaces a learned discriminator with a closed-form Fisher critic computed from frozen-teacher feature statistics.

    \item Across multiple distillation frameworks~\cite{dmd2,1_x_distill,SiD-Dit}, FisherDistill consistently improves the distilled generators and outperforms adversarial refinement in most settings, while requiring substantially fewer refinement samples.
\end{enumerate}

\section{Related Work}
\label{sec:related_work}

\paragraph{Few-step diffusion and flow distillation.}
Reducing the inference cost of diffusion and flow models has motivated many few-step distillation methods. Consistency-based approaches learn trajectory-consistent mappings~\cite{song2023consistency,ctm}, while rCM~\cite{rCM} adds distribution-matching regularization to improve fidelity. Distribution-matching methods align the student and target distributions using score-estimated KL objectives~\cite{diff_instruct,dmd} or the Fisher divergence~\cite{SiD}, or match intermediate trajectory distributions~\cite{tdm}. These ideas also extend to flow-based generators~\cite{self_flow,cai2026shortcutting,sabour2026align,SiD-Dit,senseflow}.  
Despite these advances, aggressive distillation can leave a quality gap relative to multi-step generation~\cite{amd}.

\paragraph{Adversarial refinement.}
Adversarial objectives further improve few-step generators. ADD~\cite{add} combines score distillation with an adversarial objective built on a frozen pretrained feature extractor and trainable discriminator heads, while LADD~\cite{ladd} uses generative features of a pretrained latent diffusion model. SDXL-Lightning~\cite{progressive_distill} combines adversarial training with progressive distillation, and related approaches adversarially train or post-train diffusion models for one-step image~\cite{ufogen} and video~\cite{apt} generation. %
Such objectives have also been incorporated into distribution-matching~\cite{dmd2} and score-identity~\cite{SiDA} distillation, while 1.x-Distill~\cite{1_x_distill} follows distribution matching with an adversarial stage. Although effective, these approaches rely on a learned discriminator and min-max optimization, which adds optimization and loss-balancing complexity~\cite{dmd2,SiDA,1_x_distill, twinflow, senseflow}.

\paragraph{Alternatives to adversarial refinement.}
RealUID~\cite{realuid} incorporates real-data supervision without a GAN discriminator, but retains a learned auxiliary matching function within min-max optimization. TwinFlow~\cite{twinflow} avoids an external discriminator by modifying the distillation framework to learn self-adversarial trajectories within a single model. In contrast, FisherDistill is an auxiliary refinement objective that derives its direction in closed form from real and generated statistics of frozen teacher features (\cref{fig:methods}).

\paragraph{Feature-space and statistical discrimination.}
Feature matching~\cite{feature_matching_gan} trains generators by matching intermediate discriminator features, while Fisher GAN~\cite{fisher_gan} formulates a Fisher IPM with a second-moment constraint on a learned critic. Projected GAN~\cite{projected_gan} and Vision-aided GAN~\cite{vision_aided_gan} train discriminators on fixed pretrained representations. Scale-wise Distillation~\cite{swd} introduces an MMD-based objective on frozen diffusion-model features, which reduces to feature-mean matching under a linear kernel. Our Mean variant (\cref{tab:direction_ablation}) evaluates this mean-difference direction under our logistic loss and performs worse than FisherDistill, which additionally exploits the full within-class covariance. 
More closely related, Fréchet-based objectives~\cite{fastfid,fd_loss} match the means and covariances of real and generated features in fixed pretrained representations. In contrast, we reuse intermediate representations of the frozen teacher being distilled and analytically construct a covariance-aware Fisher direction that maximizes real-vs.-generated separation relative to within-class variation, rather than minimizing a symmetric distance between distributions. 

\section{Method}
\begin{figure*}[t]
    \centering 
    \includegraphics[width=\textwidth, height=7cm]{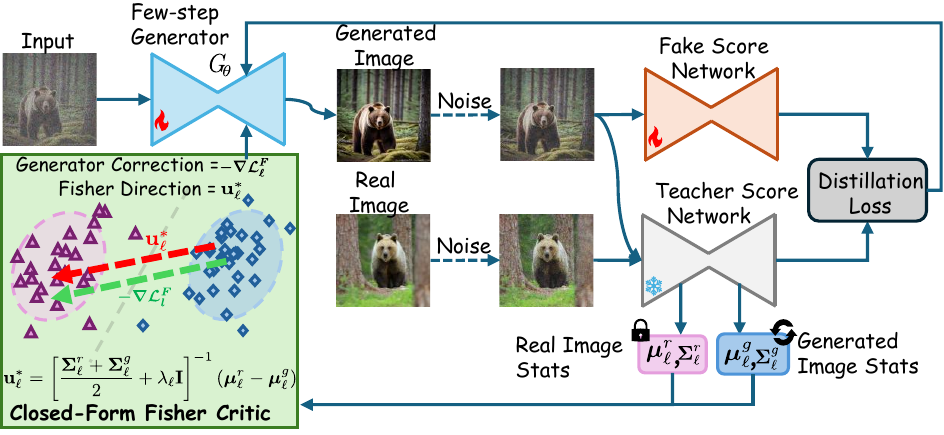}
    \caption{\textbf{Overview of FisherDistill.} We augment an existing distillation framework with Fisher refinement without modifying its base distillation objective. Real-image teacher features are precomputed offline, while generated-image features and their statistics are estimated online. Here, $\boldsymbol{\mu}_{\ell}^{r}$, $\boldsymbol{\mu}_{\ell}^{g}$
    and $\boldsymbol{\Sigma}_{\ell}^r$, $\boldsymbol{\Sigma}_{\ell}^g$ denote the means and covariances of the real and generated pooled features, respectively. At each selected layer, the closed-form Fisher direction $\mathbf{u}^{*}_{\ell}$
points from the generated toward the real feature mean, where $\lambda_{\ell}$ regularizes the covariance for numerical stability. The Fisher loss converts this direction into a corrective signal toward the real distribution, while gradients propagate through the frozen teacher to update only the generator. No learned discriminator or adversarial optimization is required.}
    \label{fig:methods}
\end{figure*}

We augment the original distribution-matching distillation objective with a critic constructed from statistics of intermediate representations of the frozen generative teacher (\cref{fig:methods}). We first describe the underlying distillation objective in a general form.
Let $\mathbf{x}^{g}=G_{\theta}(\mathbf{z}),
    \mathbf{z}\sim p_{\mathrm{noise}}$,
denote a sample from the student distribution $p_{\theta}$. Rather than
directly matching the clean, high-dimensional student and teacher
distributions, distribution-matching distillation compares their diffused
marginals. We perturb the generated sample at timestep $t$ as
\begin{equation}
    \mathbf{x}_{t}^{g}
    =
    \alpha_t \mathbf{x}^{g}
    +
    \sigma_t \boldsymbol{\epsilon},
    \qquad
    t\sim\pi(t),
    \quad
    \boldsymbol{\epsilon}\sim\mathcal{N}(\mathbf{0},\mathbf{I}),
\label{eq:noised_generated_sample}
\end{equation}
and denote the resulting student marginal by $p_{\theta}^{t}$ and the
corresponding teacher marginal by $p_{\mathrm{teacher}}^{t}$. A broad class of distribution-matching distillation objectives can then be written as
\begin{equation}
    \mathcal{L}_{\mathrm{distill}}(\theta)
    =
    \mathbb{E}_{t}
    \left[
        \mathcal{D}
        \left(
            p_{\theta}^{t}
            \,\|\, 
            p_{\mathrm{teacher}}^{t}
        \right)
    \right],
\label{eq:generic_distill}
\end{equation}
where $\mathcal{D}$ is a divergence between the diffused student and teacher marginals.
Different distillation methods correspond to different choices of $\mathcal{D}$. For example, DMD~\cite{dmd,dmd2} minimizes the reverse KL divergence $\mathbb{E}_{p_{\theta}^{t}}[\log (p_{\theta}^{t}/p_{\mathrm{teacher}}^{t})]$, whereas SiD~\cite{SiD,SiDA} minimizes the Fisher divergence $\mathbb{E}_{p_{\theta}^{t}}\|\nabla\log p_{\theta}^{t}-\nabla\log p_{\mathrm{teacher}}^{t}\|_2^2$. In practice, these objectives are optimized through method-specific estimators of the teacher and student scores.

\subsection{Role of Adversarial Refinement}
\label{subsec:role_of_adv_ref}
Adversarial refinement supplements the base distribution-matching loss in~\cref{eq:generic_distill} with an additional discriminator-derived correction
$\mathcal{L}_{\mathrm{adv}}^{G}$. We first examine this correction before introducing its closed-form Fisher counterpart.
Let $h_{\psi}(\mathbf{x}_{t}^{g},t)$ denote the discriminator logit, and let $D_{\psi}(\mathbf{x}_{t}^{g},t)
=
\sigma\!\left(h_{\psi}(\mathbf{x}_{t}^{g},t)\right)$
denote the predicted probability that $\mathbf{x}_{t}^{g}$ is real.
The non-saturating adversarial generator loss is
\begin{equation}
    \mathcal{L}_{\mathrm{adv}}^{G}
    =
    -
    \mathbb{E}_{\mathbf{z},t,\boldsymbol{\epsilon}}
    \left[
        \log D_{\psi}(\mathbf{x}_{t}^{g},t)
    \right].
\label{eq:adversarial_generator_loss}
\end{equation} The adversarially refined generator is then optimized with
$
    \mathcal{L}_{G}
    =
    \mathcal{L}_{\mathrm{distill}}
    +
    \lambda_{\mathrm{adv}}\mathcal{L}_{\mathrm{adv}}^{G}.
$
To understand the refinement signal contributed by
$\mathcal{L}_{\mathrm{adv}}^{G}$, we examine its per-sample image-space
descent direction, which decomposes as
\begin{equation}
\begin{split}
    &{-}\nabla_{\mathbf{x}^{g}}
    \left[
        -\log D_{\psi}(\mathbf{x}_{t}^{g},t)
    \right]
    \\
    &\qquad=
    \underbrace{
        \left(
            1-D_{\psi}(\mathbf{x}_{t}^{g},t)
        \right)
    }_{\text{1}}
    \underbrace{
        \alpha_t
    }_{\text{2}}
    \underbrace{
        \nabla_{\mathbf{x}_{t}^{g}}
        h_{\psi}(\mathbf{x}_{t}^{g},t)
    }_{\text{3}}.
\end{split}
\label{eq:adv_gradient_decomposition}
\end{equation}

\cref{eq:adv_gradient_decomposition} exposes three components of the
adversarial refinement signal (by the chain rule, since $\partial\mathbf{x}^{g}_{t}/\partial\mathbf{x}^{g}=\alpha_t\mathbf{I}$).
Term~(1) is a sample-dependent weight, assigning larger
corrections to samples judged less real. Term~(2), $\alpha_t$, is the
chain-rule scaling induced by the forward process. Term~(3) is the learned
discriminative direction that increases the realness logit. Thus, the
discriminator determines both the sample-wise magnitude and direction of
the refinement, while the forward process contributes the known
diffusion-dependent scaling. FisherDistill retains term~(2) but replaces
the learned discriminator-dependent terms~(1) and~(3) with their
closed-form counterparts derived from real and generated statistics in the
frozen teacher feature space.

\subsection{Residual Distribution Gaps in Teacher Features}
\label{sec:teacher_feature_gap}

The decomposition in \cref{eq:adv_gradient_decomposition} shows that the
discriminator determines which generated samples require stronger correction
and provides the direction of that correction. We ask whether this information is already present in the representations of the frozen teacher.

For the Fisher critic, we evaluate teacher features at a fixed noise level $t_F$, selected based on real-vs.-generated feature separability (more details on the selection in the supplementary). This noise level is independent of the timestep sampled by the base distillation objective. Given a clean sample $\mathbf{x}$, we construct
\begin{equation}
    \mathbf{x}_{t_F}
    =
    \alpha_{t_F}\mathbf{x}
    +
    \sigma_{t_F}\boldsymbol{\epsilon},
    \qquad
    \boldsymbol{\epsilon}\sim\mathcal{N}(\mathbf{0},\mathbf{I}),
\label{eq:fisher_noising}
\end{equation} using a fresh noise draw. For each selected teacher layer $\ell$, we denote by $\mathbf{f}_{\ell}^{s}$ the feature vector obtained from the activations of $\mathbf{x}_{t_F}^{s}$, where $s\in\{r,g\}$ denotes real and generated samples, respectively. The feature vector concatenates the spatial mean and standard deviation of each channel for U-Nets, or the corresponding statistics over image tokens for DiTs. The mean captures the average feature response, while the standard deviation preserves spatial variation such as contrast and texture. Each feature is obtained from a separate conditional teacher forward pass using the sample's own class label or prompt, without classifier-free guidance.

We measure real-vs.-generated separability at each teacher layer using the held-out AUROC of a Fisher critic. As shown in \cref{fig:layer_separability_ablation}, separability varies substantially across layers, indicating that residual distribution gaps are reflected in the frozen teacher features. We therefore construct the refinement signal directly in this space from the most separable layers and next derive the closed-form Fisher direction.

\subsection{Binary Closed-Form Fisher Critic}
\label{sec:fisher_discriminant}
Let $\boldsymbol{\mu}^{r}_{\ell},\boldsymbol{\mu}^{g}_{\ell}$ be the pooled real and generated feature means across samples at the fixed Fisher noise level $t_F$, and let $\boldsymbol{\Delta}_{\ell}$ denote their difference, i.e.,
\begin{equation}
    \boldsymbol{\mu}^{r}_{\ell}
    =
    \mathbb{E}\!\left[\mathbf{f}^{r}_{\ell}\right],
    \qquad
    \boldsymbol{\mu}^{g}_{\ell}
    =
    \mathbb{E}\!\left[\mathbf{f}^{g}_{\ell}\right],
    \qquad
    \boldsymbol{\Delta}_{\ell}
    =
    \boldsymbol{\mu}^{r}_{\ell}
    -
    \boldsymbol{\mu}^{g}_{\ell}.
\end{equation}
For this binary problem, the between-class (real-vs.-generated) scatter is $S^{B}_{\ell}=\boldsymbol{\Delta}_{\ell}\boldsymbol{\Delta}_{\ell}^{\top}$.
Let $\boldsymbol{\Sigma}^{r}_{\ell}$ and $\boldsymbol{\Sigma}^{g}_{\ell}$ denote the real and generated feature covariances, and define the shared covariance as
$
    \boldsymbol{\Sigma}_{\ell}
    =
    \frac{1}{2}
    \left(
        \boldsymbol{\Sigma}^{r}_{\ell}
        +
        \boldsymbol{\Sigma}^{g}_{\ell}
    \right)$. Fisher discriminant analysis seeks a direction that maximizes the separation between the projected class means relative to their within-class variation:
\begin{equation}
    \mathbf{u}_{\ell}^{*}
    =
    \arg\max_{\mathbf{u}}
    \frac{
        \mathbf{u}^{\top}S^{B}_{\ell}\mathbf{u}
    }{
        \mathbf{u}^{\top}\boldsymbol{\Sigma}_{\ell}\mathbf{u}
    }.
\label{eq:fisher_objective}
\end{equation}
Since $S^{B}_{\ell}=\boldsymbol{\Delta}_{\ell}\boldsymbol{\Delta}_{\ell}^{\top}$ has rank one, the binary problem has a single non-zero discriminative direction. The solution satisfies the generalized eigenvalue problem $
    S^{B}_{\ell}\mathbf{u}_{\ell}
    =
    \gamma\,
    \boldsymbol{\Sigma}_{\ell}\mathbf{u}_{\ell}$,
where $\gamma$ is the generalized eigenvalue; the derivation is provided in the supplementary material. For the binary case, this yields $\mathbf{u}_{\ell}^{*}\propto\boldsymbol{\Sigma}_{\ell}^{-1}\boldsymbol{\Delta}_{\ell}$. In practice, we regularize the covariance and use
\begin{equation}
    \mathbf{u}_{\ell}^{*}
    =
    \bar{\boldsymbol{\Sigma}}_{\ell}^{-1}
    \boldsymbol{\Delta}_{\ell},
\label{eq:fisher_direction}
\end{equation}
where $\bar{\boldsymbol{\Sigma}}_{\ell}=\boldsymbol{\Sigma}_{\ell}+\lambda_{\ell}\mathbf{I}$, with $\lambda_{\ell}>0$, is the regularized shared covariance that improves conditioning. Rather than explicitly computing
$\bar{\boldsymbol{\Sigma}}_{\ell}^{-1}$, we solve the linear system $\bar{\boldsymbol{\Sigma}}_{\ell}\mathbf{u}_{\ell}^{*}=\boldsymbol{\Delta}_{\ell}$ using a Cholesky factorization~\cite{matrix_computation}.

The Fisher direction in \cref{eq:fisher_direction} provides a one-dimensional projection along which the real and generated feature distributions are maximally separated relative to their within-class variation. We define the signed Fisher margin of a feature $\mathbf{f}$ relative to the midpoint of the projected class means as
\begin{equation}
    r_{\ell}(\mathbf{f})
    =
    \mathbf{f}^{\top}\mathbf{u}_{\ell}^{*}
    -
    \frac{1}{2}
    \left(
        {\mathbf{u}_{\ell}^{*}}^{\top}\boldsymbol{\mu}_{\ell}^{r}
        +
        {\mathbf{u}_{\ell}^{*}}^{\top}\boldsymbol{\mu}_{\ell}^{g}
    \right).
\label{eq:fisher_margin}
\end{equation}
Positive margins correspond to the real side and negative margins to the generated side.

Under the shared-covariance Gaussian model
$p(\mathbf{f}\mid s)=
\mathcal{N}(\mathbf{f};\boldsymbol{\mu}^{s}_{\ell},
\bar{\boldsymbol{\Sigma}}_{\ell})$, $s\in\{r,g\}$, the real-to-generated log-density ratio is
\begin{align}
    \log
    \frac{p(\mathbf{f}\mid r)}
         {p(\mathbf{f}\mid g)}
    &=
    \mathbf{f}^{\top}
    \bar{\boldsymbol{\Sigma}}_{\ell}^{-1}
    \boldsymbol{\Delta}_{\ell}
    -
    \frac{1}{2}
    \left(
        \boldsymbol{\mu}^{r}_{\ell}
        +
        \boldsymbol{\mu}^{g}_{\ell}
    \right)^{\top}
    \bar{\boldsymbol{\Sigma}}_{\ell}^{-1}
    \boldsymbol{\Delta}_{\ell}
    \nonumber\\
    &=
    r_{\ell}(\mathbf{f}).
\label{eq:lda_interpretation}
\end{align}
Thus, under this model, the Fisher margin is exactly the real-to-generated log-density ratio~\cite{hastie2009elements}. With equal class priors, the posterior probability that $\mathbf{f}$ is real is therefore $D^{F}_{\ell}(\mathbf{f})=\sigma(r_{\ell}(\mathbf{f}))$, a closed-form counterpart of the discriminator $D_{\psi}$ in \cref{eq:adversarial_generator_loss}. Our method,
however, constructs $r_{\ell}(\mathbf{f})$ directly from the Fisher direction and projected class means, and therefore does not require the teacher feature distributions themselves to be Gaussian.

Treating $D^{F}_{\ell}$ as a closed-form discriminator, we train the generator with the logistic loss, i.e., the non-saturating loss in \cref{eq:adversarial_generator_loss} with $D^{F}_{\ell}$ in place of $D_{\psi}$,
\begin{equation}
    \mathcal{L}^{F}_{\ell}
    =
    -\mathbb{E}
    \left[
        \log D^{F}_{\ell}(\mathbf{f}^{g}_{\ell})
    \right].
\label{eq:fisher_generator_loss}
\end{equation}
Minimizing this objective pushes generated features toward the real side of the boundary.

\subsection{Relation to Adversarial Refinement}
The Fisher critic induces a refinement direction with the same structure as
the adversarial refinement in \cref{eq:adv_gradient_decomposition}. For a
generated sample, the per-layer descent direction is
\begin{equation}
\begin{split}
    &{-}\nabla_{\mathbf{x}^{g}}
    \left[
        -\log D^{F}_{\ell}(\mathbf{f}^{g}_{\ell})
    \right]
    \\
    &\qquad=
    \underbrace{
        \left(
            1-D^{F}_{\ell}(\mathbf{f}^{g}_{\ell})
        \right)
    }_{\text{1}}
    \underbrace{
        \alpha_{t_F}
    }_{\text{2}}
    \underbrace{
        \left(
            \frac{\partial \mathbf{f}^{g}_{\ell}}
                 {\partial \mathbf{x}^{g}_{t_F}}
        \right)^{\!\top}
        \mathbf{u}^{*}_{\ell}
    }_{\text{3}}.
\end{split}
\label{eq:fisher_gradient_decomposition}
\end{equation}
Here, term~(1) is the sample-dependent weight of \cref{eq:adv_gradient_decomposition}
with the closed-form $D^{F}_{\ell}$ in place of $D_{\psi}$, term~(2) retains the
diffusion scaling at the fixed Fisher noise level, and term~(3) provides the
corrective direction by propagating the closed-form Fisher direction through
the frozen teacher. Thus, FisherDistill replaces the learned
discriminator-dependent terms with quantities obtained directly from teacher
feature statistics, without adversarial critic optimization. We derive \cref{eq:fisher_gradient_decomposition} in the supplementary material.

\subsection{Online Statistics and Lagged Fisher Updates}
\label{sec:online_fisher}

The real-data statistics are estimated once using the frozen teacher and remain
fixed throughout training. Given real features
$\{\mathbf{f}^{r}_{\ell,i}\}_{i=1}^{N_r}$, we compute
\begin{equation}
    \boldsymbol{\mu}^{r}_{\ell}
    =
    \frac{1}{N_r}\sum_{i=1}^{N_r}\mathbf{f}^{r}_{\ell,i},
    \;\;
    \boldsymbol{\Sigma}^{r}_{\ell}
    =
    \frac{1}{N_r}\sum_{i=1}^{N_r}
    \mathbf{f}^{r}_{\ell,i}\mathbf{f}^{r\top}_{\ell,i}
    -
    \boldsymbol{\mu}^{r}_{\ell}
    \boldsymbol{\mu}^{r\top}_{\ell}.
\end{equation}

Since the generated distribution evolves with $G_{\theta}$, its statistics are
tracked online. At iteration $k$, we compute the batch first and second moments
\begin{align}
    \hat{\boldsymbol{\mu}}^{g}_{\ell,k}
    &=
    \frac{1}{N_k}\sum_{i=1}^{N_k}\mathbf{f}^{g}_{\ell,i},
    &
    \hat{\mathbf{M}}^{g}_{\ell,k}
    &=
    \frac{1}{N_k}\sum_{i=1}^{N_k}
    \mathbf{f}^{g}_{\ell,i}\mathbf{f}^{g\top}_{\ell,i}.
\end{align}
We maintain exponential moving averages (EMAs)
\begin{align}
    \boldsymbol{\mu}^{g}_{\ell,k}
    &=
    \rho_k\boldsymbol{\mu}^{g}_{\ell,k-1}
    +(1-\rho_k)\hat{\boldsymbol{\mu}}^{g}_{\ell,k},
    \\
    \mathbf{M}^{g}_{\ell,k}
    &=
    \rho_k\mathbf{M}^{g}_{\ell,k-1}
    +(1-\rho_k)\hat{\mathbf{M}}^{g}_{\ell,k},
\end{align}
where $\rho_k\in[0,1)$ is the EMA decay rate, and recover the generated
covariance as
\begin{equation}
    \boldsymbol{\Sigma}^{g}_{\ell,k}
    =
    \mathbf{M}^{g}_{\ell,k}
    -
    \boldsymbol{\mu}^{g}_{\ell,k}
    \boldsymbol{\mu}^{g\top}_{\ell,k}.
\end{equation}

We use a one-step-lagged critic: at iteration $k$, the Fisher critic is constructed from fixed real statistics and generated statistics accumulated through iteration $k-1$, then held fixed while scoring the current batch. The current generated statistics are incorporated into the EMA afterward and used at iteration $k+1$. Statistics and the Fisher solve are treated as stop-gradient quantities, while gradients propagate through the current teacher features to the generator. The EMA and covariance regularization stabilize the solve, while the lag prevents the critic from being estimated from the same features it scores, which improves optimization stability, as shown in \cref{tab:fisher-impl-ablation}.

\subsection{Final Training Objective}
\label{sec:overall_objective}

For the selected teacher layers $\mathcal{S}$, we average the layer-wise Fisher losses:
\begin{equation}
    \mathcal{L}^{F}
    =
    \frac{1}{|\mathcal{S}|}
    \sum_{\ell\in\mathcal{S}}
    \mathcal{L}^{F}_{\ell}.
\label{eq:multilayer_fisher_loss}
\end{equation}
The final generator objective augments the original distillation loss with the Fisher refinement term,
\begin{equation}
    \mathcal{L}_{G}
    =
    \mathcal{L}_{\mathrm{distill}}
    +
    \lambda_{F}(k)\mathcal{L}^{F},
\label{eq:overall_generator_objective}
\end{equation}
where $\lambda_{F}(k)$ controls the refinement strength. We initially disable the Fisher loss, allowing the online generated statistics
to stabilize, and then activate it at full weight. The generator's training
schedule follows that of the base distillation framework's adversarial variant: for SiD~\cite{SiD,SiDA}, SiD-DiT~\cite{SiD-Dit} and 1.x-Distill~\cite{1_x_distill}, the generator is frozen for the first few steps, whereas for DMD2~\cite{dmd2} it is trained from the first step with the Fisher term disabled.

The teacher remains frozen throughout training, and the Fisher critics introduce no learnable parameters or separate optimizer. The generator is therefore optimized jointly by the distillation objective and the closed-form Fisher refinement.

\section{Experiments}
\label{sec:experiments}

We evaluate whether closed-form Fisher refinement can improve
few-step diffusion distillation without a learned discriminator.
We apply it to different distillation objectives on class-conditioned and text-to-image generation, and compare against the distilled baselines and their adversarially refined
variants.

\paragraph{Evaluation metrics.}
For class-conditioned generation, we report Fr\'echet Inception Distance
(FID)~\cite{fid} with precision and recall to measure
sample fidelity and coverage. For text-to-image generation, we
use FID for distributional quality, CLIP
score~\cite{clipscore} for text-image alignment,
GenEval~\cite{geneval} for compositional prompt adherence,
Aesthetic Score~\cite{aesthetics_score} for aesthetic quality, and
HPSv2~\cite{hpsv2}, ImageReward~\cite{imgrwrd}, and
PickScore~\cite{pick-a-pic} for human-preference alignment.

\subsection{Class-Conditioned Generation}
\label{sec:imagenet_results}

\paragraph{Setup.}
We evaluate class-conditioned generation on ImageNet $64\times64$ with the pretrained EDM~\cite{edm} teacher. We study Fisher refinement on two one-step distillation families: distribution matching (DMD~\cite{dmd}) and score-identity distillation (SiD~\cite{SiD}). We compare with their GAN refinements DMD2~\cite{dmd2} and SiD$^{2}$A~\cite{SiDA}, and with RealUID~\cite{realuid}, a GAN-free baseline that brings real data into the fake-score objective. For DMD, we resume from the DMD-only checkpoint (405.5K steps, FID~2.64) and fine-tune for 40K steps. For SiD, we initialize from the released checkpoint ($\alpha{=}1.2$, 939K) and follow the SiDA training budget of 2.5M generated images. RealUID uses the same initialization and budget as each Fisher model. Fisher is applied to five teacher layers for DMD2 and three for SiD. In both settings, real feature statistics are computed once from 50K images and kept fixed, while generated statistics are tracked with an EMA decay rate of $\rho_k=0.99$.

\begin{table}[t]
\centering
\caption{
\textbf{Class-conditioned ImageNet $64\times64$ results} using a pretrained EDM teacher with diffusion solvers and prior distillation methods. FisherDistill is evaluated on top of DMD2 and SiD, with precision and recall reported where available. \textbf{Bold}: best among the compared methods below the faint rule within each distribution-matching block. Results outside the ``Critic Based'' family are from~\cite{SiD} or original papers. $^{\dagger}$: reproduced with the official code under a matched training budget.
}
\label{tab:imagenet64_label}

\setlength{\tabcolsep}{3.2pt}
\renewcommand{\arraystretch}{1.02}
\scriptsize
\resizebox{\columnwidth}{!}{%
\begin{tabular}{@{}llcccc@{}}
\toprule
\textbf{Family}
& \textbf{Model}
& \textbf{NFE}
& \textbf{FID} $\downarrow$
& \textbf{Prec.} $\uparrow$
& \textbf{Rec.} $\uparrow$ \\
\midrule

\multirow{2}{*}{Teacher}
& VP EDM ODE
& 511 & 2.32 &  &  \\

& VP EDM SDE
& 511 & 1.32 &  &  \\
\midrule

\multirow{5}{*}{Solver}
& DDIM~\citep{ddim}
& 8 & 22.59 &  &  \\

& DPM-Solver~\citep{dpm}
& 8 & 12.42 &  &  \\

& AMED~\citep{zhou2024fast}
& 8 & 7.01 &  &  \\

& GITS~\citep{chen2024trajectory}
& 8 & 5.64 &  &  \\

& 
INDIS~\citep{INDIS}
& 8 & 4.49 &  &  \\
\midrule

\multirow{8}{*}{Distillation}
& PD~\cite{progressive_distill}
& 2 & 8.95 & 0.63 & 0.65 \\

& CD-LPIPS~\cite{song2023consistency}
& 2 & 4.70 & 0.69 & 0.64 \\

& DFNO~\cite{dfno}
& 1 & 7.83 &  & 0.61 \\

& Diff-Instruct~\cite{diff_instruct}
& 1 & 5.57 &  &  \\

& iCT~\cite{ict}
& 1 & 4.02 & 0.70 & 0.63 \\

& iCT-deep~\cite{ict}
& 1 & 3.25 & 0.72 & 0.63 \\

& EMD-16~\cite{em_distill}
& 1 & 2.20 &  &  \\

& CTM~\cite{ctm}
& 1 & 1.92 &  & 0.57 \\

\midrule

\multirow{8}{*}{Critic Based}
& DMD~\cite{dmd}
& 1 & 2.64 & 0.76 & 0.59 \\
\arrayrulecolor{black!25}
\cmidrule(lr){2-6}
\arrayrulecolor{black}
& DMD2$^{\dagger}$~\cite{dmd2}
& 1 & 1.42 & \textbf{0.74} & 0.60 \\

& RealUID$^{\dagger}$~\cite{realuid}
& 1 & 3.07 & \textbf{0.74} & 0.58 \\

& \cellcolor{green!25}\textbf{FisherDistill}
& \cellcolor{green!25}1 & \cellcolor{green!25}\textbf{1.40} & \cellcolor{green!25}\textbf{0.74} & \cellcolor{green!25}\textbf{0.62} \\

\arrayrulecolor{black!100}
\cmidrule(lr){2-6}
\arrayrulecolor{black}

& SiD~\cite{SiD}, $\alpha=1.2$
& 1 & 1.54 & 0.74 & 0.63 \\
\arrayrulecolor{black!25}
\cmidrule(lr){2-6}
\arrayrulecolor{black}
& SiD$^{2}$A$^{\dagger}$~\cite{SiDA}, $\alpha=1.0$
& 1 & \textbf{1.31} & 0.74 & 0.62 \\
&
RealUID$^{\dagger}$~\cite{realuid}
& 1 & 1.45 & 0.74 & 0.62 \\

& \cellcolor{green!25}\textbf{FisherDistill}
& \cellcolor{green!25}1 & \cellcolor{green!25}1.34 & \cellcolor{green!25}\textbf{0.76} & \cellcolor{green!25}\textbf{0.63} \\

\bottomrule
\end{tabular}%
}
\end{table}
\paragraph{Results.}
\cref{tab:imagenet64_label} shows that Fisher refinement consistently improves both distillation objectives. On DMD2, Fisher reduces FID from $1.42$ to $1.40$ and improves recall from $0.60$ to $0.62$, while preserving precision at $0.74$. It also outperforms RealUID, with substantially lower FID ($1.40$ vs.\ $3.07$) and higher recall ($0.62$ vs.\ $0.58$) at the same precision. On SiD, Fisher reduces FID from $1.54$ to $1.34$ and increases precision from $0.74$ to $0.76$ without reducing recall. Compared with RealUID, Fisher achieves lower FID and higher precision and recall. Fisher is also competitive with adversarially refined SiD$^{2}$A. Overall, Fisher refinement improves sample fidelity without compromising distributional coverage. Fisher also needs fewer refinement samples than GAN refinement: on DMD2 it reaches its reported FID after 6.24M samples versus 12.64M for the matched GAN arm ($2.0\times$ fewer samples, $2.7\times$ fewer GPU-hours), and on SiD it reaches its best FID after 1.48M samples, a level SiDA reaches only after 1.97M ($1.3\times$ fewer). 

\subsection{Text-to-Image Generation}
\label{sec:t2i_results}
\paragraph{Setup.}
We evaluate Fisher refinement in three text-to-image settings: DMD2~\cite{dmd2} with SD~v1.5~\cite{ldm}, 1.x-Distill~\cite{1_x_distill} with
SD3-Medium~\cite{sd3}, and SiD-DiT~\cite{SiD-Dit} with SANA-0.6B and SANA-1.6B~\cite{sana}. We follow the baseline protocols and compare with the matched GAN refinement and GAN-free RealUID~\cite{realuid}, starting from the same stage-1 checkpoints, i.e., models before GAN fine-tuning.

For DMD2, we fine-tune the released DMD-only checkpoint on LAION-Aesthetics prompts~\cite{laion}. Fisher uses three frozen teacher layers, with real statistics precomputed from 8K random images from DMD2's GAN training set.

For 1.x-Distill, whose official weights are unavailable, we reproduce the Stage-I student and 8-step reference model using 20K Midjourney prompts~\cite{midjourneyv6llava}, followed by Stage-II refinement for 2K iterations. Fisher uses three frozen teacher blocks, with real statistics precomputed from 10K latents generated by SD3-Medium (28 steps, CFG 7.0). Both the GAN baseline and RealUID use the 8-step reference model as real data. RealUID replaces the Stage-II GAN loss, and all methods use a matched generator-update budget.

For SiD-DiT, we refine the released SANA checkpoints at both scales using LAION-Aesthetics prompts~\cite{laion}. Fisher uses ten (0.6B) and eight (1.6B) frozen teacher layers, with real statistics precomputed from 10K Midjourney images~\cite{midjourneyv6llava} also used by the SiD$^{2}$A discriminator and RealUID. All models are evaluated zero-shot on COCO-2014~\cite{coco}.

\begin{figure*}[t]
    \centering
    \includegraphics[width=\textwidth]{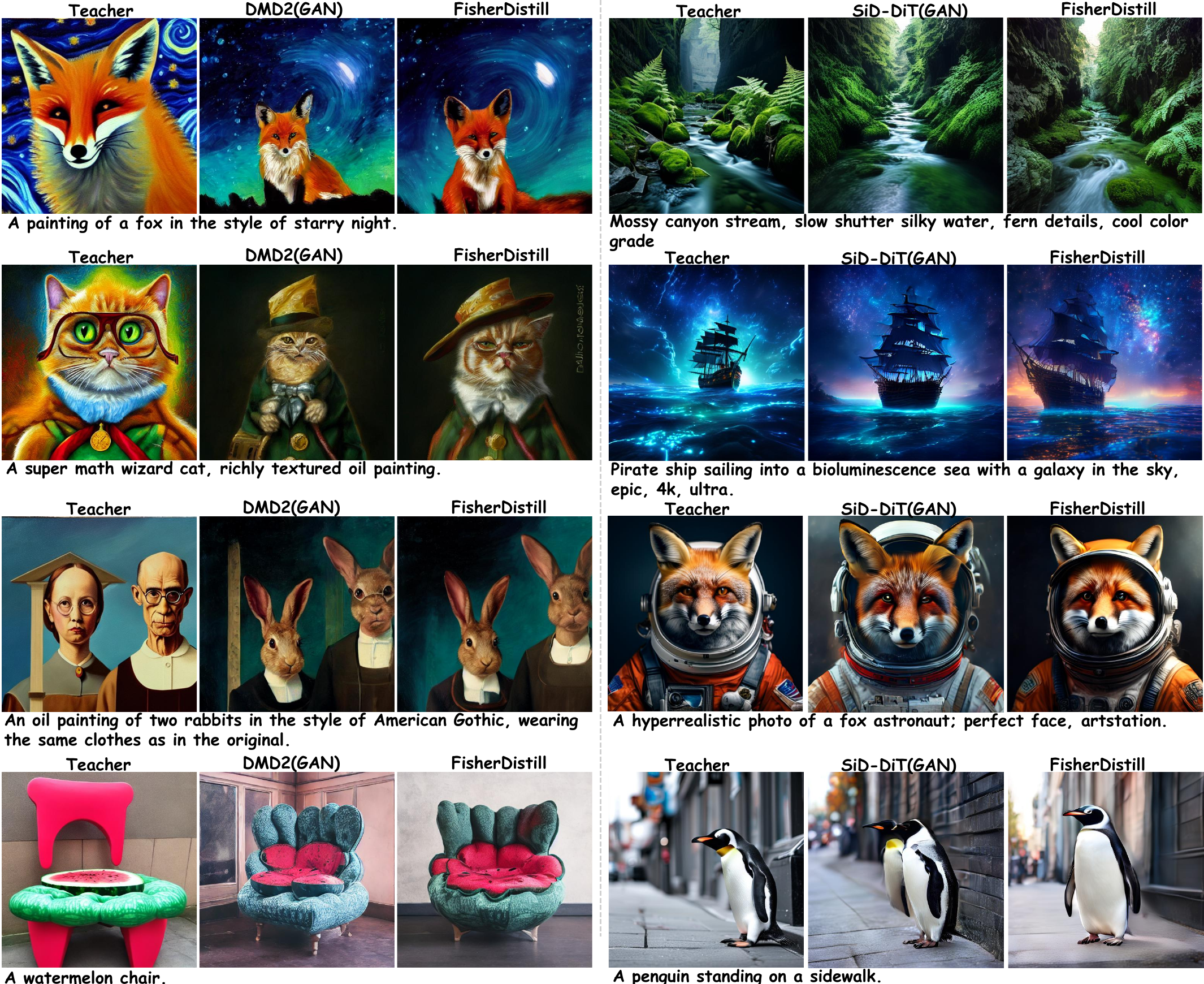}
    \caption{\textbf{512$\times$512 samples produced by our 1-step generators} distilled from SD~v1.5 (left) and SANA-1.6B (right), compared with the teachers and the GAN-refined DMD2 and SiD-DiT. 
    }
    \label{fig:qual_t2i_comparison}
\end{figure*}

\begin{table*}[t]
\centering
\caption{
\textbf{Text-to-image generation} across distillation frameworks and teacher models.
We compare Fisher refinement with adversarial refinement and RealUID.
\textbf{Bold} and \underline{underline} denote the best and second-best results, respectively,
among the three refinement methods within each setting; ties receive the same formatting.
$\dagger$: trained by us with the official code under a matched training budget. $\times$2 indicates classifier-free guidance.
}
\label{tab:t2i_comparison}

\setlength{\tabcolsep}{4.2pt}
\renewcommand{\arraystretch}{1.05}
\scriptsize
\resizebox{0.95\textwidth}{!}{%
\begin{tabular}{@{}lcccccccc@{}}
\toprule
\textbf{Model}
& \textbf{NFE}
& \textbf{FID} $\downarrow$
& \textbf{CLIP} $\uparrow$
& \textbf{GenEval} $\uparrow$
& \textbf{Aesth.} $\uparrow$
& \textbf{HPSv2} $\uparrow$
& \textbf{ImgRwd} $\uparrow$
& \textbf{PickScore} $\uparrow$ \\
\midrule

\textbf{SD v1.5}~\cite{ldm}
& 50$\times$2
& 8.60
& 0.310
& 0.442
& 5.466
& 0.267
& 0.096
& 20.61 \\

DMD~\cite{dmd}
& 1
& 9.28
& 0.302
& 0.347
& 5.333
& 0.255
& -0.410
& 20.01 \\

\faintrule

$\text{DMD2}^{\dagger}$~\cite{dmd2}
& 1
& \underline{8.58}
& 0.300
& 0.343
& \textbf{5.414}
& \underline{0.256}
& -0.378
& 20.00 \\

$\text{RealUID}^{\dagger}$~\cite{realuid}
& 1
& 9.37
& \underline{0.303}
& \textbf{0.352}
& 5.346
& \underline{0.256}
& \textbf{-0.332}
& \underline{20.06} \\

\rowcolor{green!25}
\textbf{FisherDistill}
& 1
& \textbf{8.46}
& \textbf{0.304}
& \underline{0.348}
& \underline{5.393}
& \textbf{0.257}
& \underline{-0.350}
& \textbf{20.07} \\

\midrule

\textbf{SD3-Medium}~\cite{sd3}
& 28$\times$2
& 23.39
& 0.330
& 0.708
& 5.370
& 0.294
& 1.027
& 22.50 \\

$\text{1.x-Distill Stage I}^{\dagger}$~\cite{1_x_distill}
& 2
& 32.35
& 0.329
& 0.651
& 5.283
& 0.284
& 0.985
& 22.15 \\

\faintrule

$\text{1.x-Distill Stage II}^{\dagger}$~\cite{1_x_distill}
& 2
& \underline{30.20}
& \textbf{0.330}
& \underline{0.655}
& \underline{5.430}
& \underline{0.288}
& \underline{0.998}
& \textbf{22.59} \\

$\text{RealUID}^{\dagger}$~\cite{realuid}
& 2
& 34.03
& 0.326
& 0.641
& 5.101
& 0.279
& 0.862
& 21.99 \\

\rowcolor{green!25}
\textbf{FisherDistill}
& 2
& \textbf{29.54}
& \textbf{0.330}
& \textbf{0.682}
& \textbf{5.483}
& \textbf{0.294}
& \textbf{1.059}
& \underline{22.37} \\

\midrule

\textbf{SANA 0.6B}~\cite{sana}
& 20$\times$2
& 26.38
& 0.338
& 0.651
& 5.730
& 0.311
& 0.999
& 22.50 \\

SiD-DiT~\cite{SiD-Dit}
& 1
& 28.94
& 0.323
& 0.594
& 5.751
& 0.278
& 0.709
& 21.51 \\

\faintrule

$\text{SiD$_2^{a}$-DiT}^{\dagger}$~\cite{SiD-Dit}
& 1
& 34.13
& 0.298
& 0.486
& 5.383
& 0.244
& 0.287
& 21.00 \\

$\text{RealUID}^{\dagger}$~\cite{realuid}
& 1
& \underline{31.87}
& \underline{0.320}
& \underline{0.590}
& \underline{5.740}
& \underline{0.274}
& \underline{0.688}
& \underline{21.46} \\

\rowcolor{green!25}
\textbf{FisherDistill}
& 1
& \textbf{25.24}
& \textbf{0.326}
& \textbf{0.597}
& \textbf{5.744}
& \textbf{0.286}
& \textbf{0.820}
& \textbf{21.80} \\

\midrule

\textbf{SANA 1.6B}~\cite{sana}
& 20$\times$2
& 26.83
& 0.334
& 0.652
& 5.840
& 0.312
& 1.054
& 22.56 \\

SiD-DiT~\cite{SiD-Dit}
& 1
& 28.86
& 0.325
& 0.581
& 5.881
& 0.285
& 0.880
& 21.71 \\

\faintrule

$\text{SiD$_2^{a}$-DiT}^{\dagger}$~\cite{SiD-Dit}
& 1
& \underline{28.58}
& \underline{0.323}
& \textbf{0.615}
& \underline{5.889}
& \textbf{0.289}
& \underline{0.883}
& \textbf{21.81} \\

$\text{RealUID}^{\dagger}$~\cite{realuid}
& 1
& 34.28
& 0.322
& 0.578
& \textbf{5.916}
& 0.279
& 0.867
& 21.60 \\

\rowcolor{green!25}
\textbf{FisherDistill}
& 1
& \textbf{26.73}
& \textbf{0.326}
& \underline{0.603}
& 5.816
& \textbf{0.289}
& \textbf{0.884}
& \underline{21.73} \\

\bottomrule
\end{tabular}%
}
\end{table*}

\paragraph{Evaluation protocols.} Since each baseline has its own protocol, we evaluate every method with the protocol of
its base framework and compare only within each group of~\cref{tab:t2i_comparison}. Details are in the supplementary material.

\paragraph{Results.}
As shown in \cref{tab:t2i_comparison}, Fisher consistently improves distributional quality across all settings. On DMD2, FID improves from $8.58$ to $8.46$, while on 1.x-Distill it improves from $32.35/30.20$ for Stages~I/II to $29.54$, with competitive alignment and preference metrics. The same trend holds for SANA at both scales; for example, on SANA-0.6B Fisher reduces FID from $28.94$ to $25.24$ and improves most alignment and preference metrics. In contrast, RealUID does not improve the first-stage models on FID and mostly fails to improve the other metrics. Notably, the one-step Fisher models also match or surpass the multi-step SANA teachers in FID at both scales ($25.24$ vs.\ $26.38$ and $26.73$ vs.\ $26.83$). Overall, Fisher improves FID across architectures, scales, and distillation objectives while preserving competitive semantic and perceptual quality (\cref{fig:qual_t2i_comparison}).

\subsection{Ablations}
\begin{figure*}[t]
    \centering 
    \includegraphics[width=\textwidth]{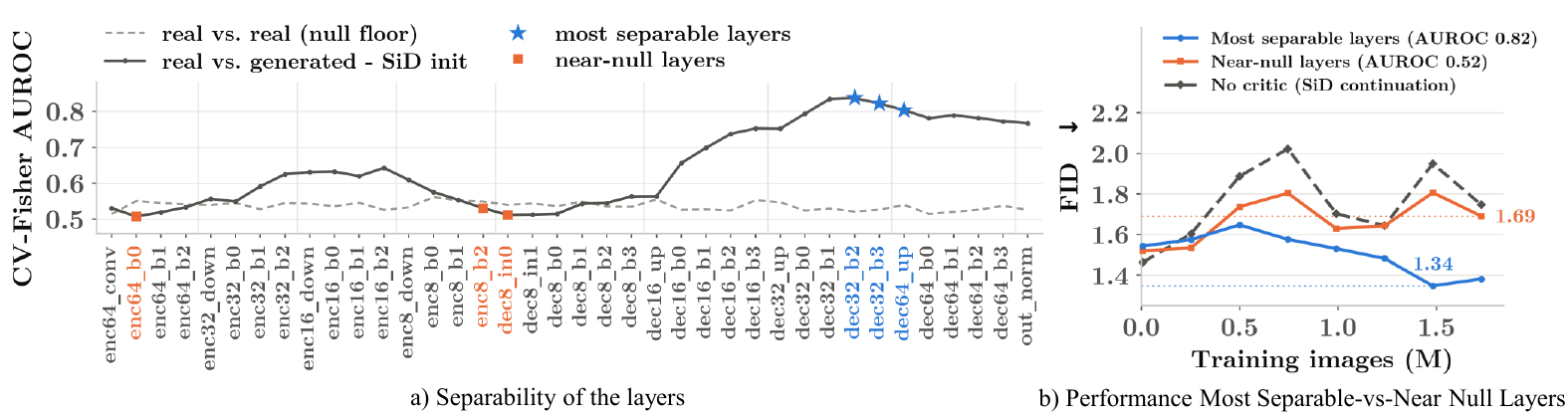}
    \caption{\textbf{Layer-separability ablation} on EDM~\cite{edm}/ImageNet $64\times64$.
    (a) Held-out Fisher AUROC across frozen teacher layers for real-vs.-generated
    features, together with a real-vs.-real control that estimates the null floor.
    Blue stars mark the three most separable layers and orange squares mark three
    near-null layers. (b) Training with the separable layers yields substantially
    better FID than near-null layers, which closely track the no-critic
    continuation.}
\label{fig:layer_separability_ablation}
\end{figure*}

\paragraph{Layer selection for Fisher discriminants.}
The closed-form Fisher critic reads generated samples through a small set of
teacher layers, so its usefulness depends on whether those layers can
distinguish real from generated features. \Cref{fig:layer_separability_ablation}(a) quantifies this: for each of the 38 pooled teacher layers, we fit a Fisher discriminant between real images and samples from the SiD initialization~\cite{SiD} on ImageNet~\cite{imagnet} and report its held-out AUROC (5-fold cross-validation). A real-vs.-real split of the dataset gives the null floor.
Separability varies substantially across layers: mid-resolution decoder blocks reach AUROC $0.80$--$0.84$, while several layers remain close to the floor. We select the three most separable layers (\emph{best arm}, \textcolor{blue}{\(\star\)}) and three near-null layers (\emph{null arm}, \textcolor{orange}{\(\scalebox{0.5}{$\blacksquare$}\)}, AUROC $0.51$--$0.53$ versus a floor of ${\approx}0.54$), and train both under identical settings.

As shown in \cref{fig:layer_separability_ablation}(b), the separable layers
yield a stable improvement to an FID of $1.34$, while the near-null layers reach only
$1.69$ and behave similarly to continuing training without the critic.
This result shows that Fisher refinement is effective only when the selected teacher representation contains reliable real-vs.-generated information. The held-out
AUROC sweep therefore provides a simple criterion for selecting layers before
training.

\paragraph{Effect of Fisher refinement.}
Since the Fisher critic is introduced while fine-tuning an already distilled generator, improvements could arise simply from additional score-distillation updates. We therefore compare Fisher against continuing pure SiD from the same initialization under a matched training budget. As shown in \cref{tab:matched_anchor_nfe1}, Fisher improves single-step FID on ImageNet-64 from $1.75$ to $1.38$. On SANA-0.6B~\cite{sana}, FID improves from $28.45$ to $27.33$, with CLIP unchanged at $0.33$. These results indicate that the gains arise from the additional Fisher correction rather than continued score distillation alone.

\begin{table}[t]
\centering
\caption{\textbf{Continuation of distillation objective}, NFE=1.
Starting from the same initialization, we compare continued SiD~\cite{SiD} training with and without the proposed Fisher refinement. Fisher improves FID in both EDM~\cite{edm}/ImageNet-64 and SANA-0.6B~\cite{sana}/COCO-10K at a comparable budget (1.7M images).} 

\label{tab:matched_anchor_nfe1}
\small
\setlength{\tabcolsep}{4pt}
\begin{tabular}{lccc}
\toprule
 & EDM/ImageNet-64 & \multicolumn{2}{c}{SANA-0.6B/COCO-30K} \\
\cmidrule(lr){2-2}\cmidrule(lr){3-4}
Method & FID\,$\downarrow$ & FID\,$\downarrow$ & CLIP\,$\uparrow$ \\
\midrule
SiD only & 1.75 & 28.45 & \textbf{0.33} \\
\rowcolor{green!25}
FisherDistill & \textbf{1.38} & \textbf{27.33} & \textbf{0.33} \\
\bottomrule
\end{tabular}
\end{table}

\paragraph{Effect of the full covariance.}
Simpler discriminant directions could also pull generated teacher features toward the real ones without modeling the full within-class covariance. We therefore compare the raw mean difference (Mean), per-dimension variance normalization (Diag), and the full Fisher direction (Full), while matching their initial gradient magnitudes. As shown in \cref{tab:direction_ablation}, incorporating the full covariance
is important for distributional quality. Full improves FID from $9.28$ at
initialization to $8.46$, whereas Diag reaches $9.18$ and Mean degrades FID
to $9.85$. In contrast, CLIP and preference metrics vary only modestly. These results indicate that modeling cross-feature
correlations provides information beyond the mean difference or independent
per-feature variances, particularly for FID.

\begin{table}[t]
\centering
\small
\setlength{\tabcolsep}{3.2pt}
\caption{\textbf{Fisher covariance structure ablation} on SD v1.5~\cite{ldm} DMD2~\cite{dmd2}/LAION setup, NFE=1. Mean uses only the class-mean difference, $\mathbf{u}_{\ell}=\boldsymbol{\Delta}_{\ell}$; Diag additionally accounts for per-feature variance, $\mathbf{u}_{\ell}=(\operatorname{diag}\boldsymbol{\Sigma}_{\ell}+\lambda_{\ell}\mathbf{I})^{-1}\boldsymbol{\Delta}_{\ell}$; Full uses the complete feature covariance, $\mathbf{u}_{\ell}=(\boldsymbol{\Sigma}_{\ell}+\lambda_{\ell}\mathbf{I})^{-1}\boldsymbol{\Delta}_{\ell}$, as in \cref{eq:fisher_direction}. All arms resume the same released checkpoint (Init.). }

\label{tab:direction_ablation}
\resizebox{0.95\columnwidth}{!}{%
\begin{tabular}{@{}lccccc@{}}
\toprule
Direction $\mathbf{u}_{\ell}$ & FID$\downarrow$ & CLIP$\uparrow$ & HPSv2$\uparrow$ &
ImgRwd$\uparrow$ & PickScr$\uparrow$ \\
\midrule
Init. (no refinement)
& 9.28
& 0.302
& 0.255
& $-$0.410
& 20.01 \\
\faintrule
Mean
& 9.85
& \textbf{0.305}
& \textbf{0.258}
& $-$0.394
& 20.03 \\

Diag
& 9.18
& \textbf{0.305}
& \textbf{0.258}
& $-$0.358
& \textbf{20.07} \\

\rowcolor{green!25}
Full
& \textbf{8.46}
& 0.304
& 0.257
& \textbf{$-$0.350}
& \textbf{20.07} \\
\bottomrule
\end{tabular}}
\end{table}  

\paragraph{Fisher critic life-cycle.}
All variants in \cref{tab:fisher-impl-ablation} start from the same SiD-initialized generator~\cite{SiD} (FID $1.52$) and differ only
in the critic update life-cycle and the staging of training. In the
legacy life-cycle, generated statistics are updated at every
gradient-accumulation round. With 64 accumulation rounds per optimizer step,
this changes the Fisher critic 64 times per generator update,
shortens the effective memory of the statistics EMA to roughly one step
($0.99^{64}\approx0.53$), and allows current samples to enter the
statistics used to score subsequent samples. This configuration degrades FID to $1.63$ and exhibits the strongest precision-recall trade-off in the table: precision increases by $0.031$, while recall decreases by $0.036$. This trade-off suggests that the unstable critic favors precision at the
expense of coverage.

Updating the statistics once per optimizer step and scoring the current batch
with the critic from the previous step removes both within-step drift and
self-scoring. This lagged life-cycle improves FID to $1.38$ and raises recall
from $0.594$ to $0.614$, while retaining high precision, which indicates
a more balanced fidelity--coverage trade-off than the legacy scheme. Finally, we follow SiDA's~\cite{SiDA} staged training strategy: the generator is initially
held fixed while the Fisher statistics stabilize, and Fisher
refinement is activated only afterward. We match SiDA's loss weighting so that
the two settings differ only in the refinement term. This staged activation
reduces the early optimization transient and yields our full method with FID
$1.34$. 
Together, these results show that per-step lagged updates and delayed
Fisher activation are important for stable training under gradient accumulation.

\begin{table}[t]
\centering
\small
\setlength{\tabcolsep}{3.0pt}
\renewcommand{\arraystretch}{1.03}

\caption{\textbf{Fisher critic life-cycle ablation} on EDM~\cite{edm}/ImageNet 64$\times$64, NFE=1. All variants start from the same SiD~\cite{SiD} initialization and use the same Fisher formulation under a matched 2.5M-image budget. Updating the
critic once per optimizer step with a one-step lag substantially improves FID,
while delayed Fisher activation as in SiDA~\cite{SiDA} provides a further gain. Results are mean$\pm$std
over three evaluation trials.}
\label{tab:fisher-impl-ablation}

\resizebox{\columnwidth}{!}{%
\begin{tabular}{@{}lccc@{}}
\toprule
\textbf{Configuration}
& \textbf{FID}$\downarrow$
& \textbf{Prec.}$\uparrow$
& \textbf{Rec.}$\uparrow$ \\
\midrule

Init
& 1.52 $\pm$ 0.02
& 0.740 $\pm$ 0.001
& 0.627 $\pm$ 0.001 \\

\faintrule

$+$ Legacy life-cycle
& 1.63 $\pm$ 0.05
& 0.771 $\pm$ 0.000
& 0.594 $\pm$ 0.001 \\

$+$ Per-step lagged
& 1.38 $\pm$ 0.01
& 0.754 $\pm$ 0.002
& 0.614 $\pm$ 0.005 \\

\rowcolor{green!25}
$+$ Delayed activation (Ours)
& \textbf{1.34} $\pm$ \textbf{0.02}
& \textbf{0.756} $\pm$ \textbf{0.002}
& \textbf{0.626} $\pm$ \textbf{0.004} \\

\bottomrule
\end{tabular}%
}
\end{table}

\section{Conclusion}
We introduced \textbf{FisherDistill}, a simple non-adversarial refinement method for few-step generative distillation. Rather than learning an adversarial discriminator, FisherDistill exploits intermediate representations of the frozen generative teacher and derives a closed-form Fisher critic from the first- and second-order statistics of real and generated features. The resulting covariance-aware refinement signal is incorporated alongside the original distillation objective without requiring a learned critic or alternating adversarial optimization. Across multiple distillation frameworks and generation settings, FisherDistill consistently improves the underlying distilled generators and surpasses their adversarially refined counterparts. These results suggest that the residual distribution gap left by aggressive distillation can be effectively corrected using simple statistics already exposed by the frozen teacher, providing a lightweight alternative to learned adversarial refinement.


{
    \small
    \bibliographystyle{ieeenat_fullname}
    \bibliography{main}
}

\clearpage
\appendix
\maketitlesupplementary

\section{Derivation of the Adversarial Refinement Direction}
\label{sec:adv_gradient_derivation}

We derive the per-sample descent direction used in
\cref{eq:adv_gradient_decomposition}. For a generated sample $\mathbf{x}_{t}^{g}$, the
non-saturating adversarial loss is given by,
\begin{equation}
    \ell_{\mathrm{adv}}^{G}
    =
    -\log D_{\psi}(\mathbf{x}_{t}^{g},t),
\end{equation}
where
\begin{equation}
    D_{\psi}(\mathbf{x}_{t}^{g},t)
    =
    \sigma\!\left(
        h_{\psi}(\mathbf{x}_{t}^{g},t)
    \right).
\end{equation}
For brevity, let
$D_{\psi}=D_{\psi}(\mathbf{x}_{t}^{g},t)$ and
$h_{\psi}=h_{\psi}(\mathbf{x}_{t}^{g},t)$. Since
\begin{equation}
    \frac{d}{dh}
    \left[-\log \sigma(h)\right]
    =
    -\left(1-\sigma(h)\right),
\end{equation}
the gradient with respect to the noised generated sample is
\begin{equation}
    \nabla_{\mathbf{x}_{t}^{g}}
    \ell_{\mathrm{adv}}^{G}
    =
    -\left(1-D_{\psi}\right)
    \nabla_{\mathbf{x}_{t}^{g}}h_{\psi}.
\label{eq:supp_adv_grad_xt}
\end{equation}
From the forward process, $\mathbf{x}_{t}^{g}=\alpha_t\mathbf{x}^{g} + \sigma_t\boldsymbol{\epsilon}$, we have,
\begin{equation}
    \frac{\partial \mathbf{x}_{t}^{g}}
         {\partial \mathbf{x}^{g}}
    =
    \alpha_t\mathbf{I}.
\end{equation}
Applying the chain rule gives,
\begin{align}
    \nabla_{\mathbf{x}^{g}}
    \ell_{\mathrm{adv}}^{G}
    &=
    \alpha_t
    \nabla_{\mathbf{x}_{t}^{g}}
    \ell_{\mathrm{adv}}^{G}
    \\
    &=
    -\left(1-D_{\psi}\right)
    \alpha_t
    \nabla_{\mathbf{x}_{t}^{g}}h_{\psi}.
\end{align}
Therefore, the corresponding descent direction is:
\begin{equation}
    -\nabla_{\mathbf{x}^{g}}
    \ell_{\mathrm{adv}}^{G}
    =
    \underbrace{
        \left(1-D_{\psi}(\mathbf{x}_{t}^{g},t)\right)
    }_{\text{sample weight}}
    \underbrace{\alpha_t}_{\text{diffusion scaling}}
    \underbrace{
        \nabla_{\mathbf{x}_{t}^{g}}
        h_{\psi}(\mathbf{x}_{t}^{g},t)
    }_{\text{discriminative direction}},
\end{equation}
which gives \cref{eq:adv_gradient_decomposition}.

\subsection{Generalized Eigenvalue Derivation for the Binary Fisher Critic}

We provide the derivation of the generalized eigenvalue problem used in
\cref{sec:fisher_discriminant}. The goal of Fisher discriminant analysis
is to find a projection direction along which the separation between the real
and generated feature means is large relative to the variation within the two
classes. For a projection direction $\mathbf{u}_{\ell}$, this trade-off is measured by
the Fisher criterion
\begin{equation}
    J(\mathbf{u}_{\ell})
    =
    \frac{
        \mathbf{u}_{\ell}^{\top}S^{B}_{\ell}\mathbf{u}_{\ell}
    }{
        \mathbf{u}_{\ell}^{\top}\boldsymbol{\Sigma}_{\ell}\mathbf{u}_{\ell}
    }.
\label{eq:supp_fisher_criterion}
\end{equation}
The numerator measures the separation between the projected class means,
whereas the denominator measures the corresponding within-class variation.

The value of $J(\mathbf{u}_{\ell})$ depends only on the direction of
$\mathbf{u}_{\ell}$. Indeed, for any non-zero scalar $a$,
\begin{equation}
    J(a\mathbf{u}_{\ell})
    =
    \frac{
        a^{2}\mathbf{u}_{\ell}^{\top}S^{B}_{\ell}\mathbf{u}_{\ell}
    }{
        a^{2}\mathbf{u}_{\ell}^{\top}\boldsymbol{\Sigma}_{\ell}\mathbf{u}_{\ell}
    }
    =
    J(\mathbf{u}_{\ell}).
\end{equation}
We can therefore fix the scale of $\mathbf{u}_{\ell}$ without changing the
optimization. Choosing
$\mathbf{u}_{\ell}^{\top}\boldsymbol{\Sigma}_{\ell}\mathbf{u}_{\ell}=1$
turns the Fisher criterion into the equivalent constrained problem
\begin{equation}
    \max_{\mathbf{u}_{\ell}}
    \quad
    \mathbf{u}_{\ell}^{\top}S^{B}_{\ell}\mathbf{u}_{\ell}
    \qquad
    \text{subject to}
    \qquad
    \mathbf{u}_{\ell}^{\top}\boldsymbol{\Sigma}_{\ell}\mathbf{u}_{\ell}=1.
\label{eq:supp_fisher_constrained}
\end{equation}

Introducing a Lagrange multiplier $\gamma$ for the constraint gives
\begin{equation}
    \mathcal{J}(\mathbf{u}_{\ell},\gamma)
    =
    \mathbf{u}_{\ell}^{\top}S^{B}_{\ell}\mathbf{u}_{\ell}
    -
    \gamma
    \left(
        \mathbf{u}_{\ell}^{\top}\boldsymbol{\Sigma}_{\ell}\mathbf{u}_{\ell}
        -1
    \right).
\end{equation}
At a stationary point, the derivative with respect to $\mathbf{u}_{\ell}$ must
vanish. Since both $S^{B}_{\ell}$ and
$\boldsymbol{\Sigma}_{\ell}$ are symmetric,
\begin{equation}
    \frac{\partial \mathcal{J}}
         {\partial \mathbf{u}_{\ell}}
    =
    2S^{B}_{\ell}\mathbf{u}_{\ell}
    -
    2\gamma\boldsymbol{\Sigma}_{\ell}\mathbf{u}_{\ell}
    =
    0.
\end{equation}
Rearranging gives
\begin{equation}
    S^{B}_{\ell}\mathbf{u}_{\ell}
    =
    \gamma
    \boldsymbol{\Sigma}_{\ell}\mathbf{u}_{\ell},
\label{eq:supp_generalized_eigen}
\end{equation}
which is the generalized eigenvalue problem associated with the matrix pair
$(S^{B}_{\ell},\boldsymbol{\Sigma}_{\ell})$. Multiplying both sides by
$\mathbf{u}_{\ell}^{\top}$ shows that
\begin{equation}
    \gamma
    =
    \frac{
        \mathbf{u}_{\ell}^{\top}S^{B}_{\ell}\mathbf{u}_{\ell}
    }{
        \mathbf{u}_{\ell}^{\top}\boldsymbol{\Sigma}_{\ell}\mathbf{u}_{\ell}
    }
    =
    J(\mathbf{u}_{\ell}),
\end{equation}
so the generalized eigenvalue is exactly the Fisher criterion attained by
that direction. The maximizing solution therefore corresponds to the largest
generalized eigenvalue.

For our binary real-vs.-generated problem,
\begin{equation}
    S^{B}_{\ell}
    =
    \boldsymbol{\Delta}_{\ell}
    \boldsymbol{\Delta}_{\ell}^{\top},
\end{equation}
which has rank one. Substituting this form into
\cref{eq:supp_generalized_eigen} gives
\begin{equation}
    \boldsymbol{\Delta}_{\ell}
    \left(
        \boldsymbol{\Delta}_{\ell}^{\top}\mathbf{u}_{\ell}
    \right)
    =
    \gamma
    \boldsymbol{\Sigma}_{\ell}\mathbf{u}_{\ell}.
\end{equation}
The quantity
$\boldsymbol{\Delta}_{\ell}^{\top}\mathbf{u}_{\ell}$ is a scalar, so for the
non-zero generalized eigenvalue,
$\boldsymbol{\Sigma}_{\ell}\mathbf{u}_{\ell}$ must lie in the direction of
$\boldsymbol{\Delta}_{\ell}$. When
$\boldsymbol{\Sigma}_{\ell}$ is invertible, this gives
\begin{equation}
    \mathbf{u}_{\ell}^{*}
    \propto
    \boldsymbol{\Sigma}_{\ell}^{-1}
    \boldsymbol{\Delta}_{\ell}.
\end{equation}
Thus, in the binary case, the Fisher criterion yields a single non-zero
discriminative direction given by the covariance-adjusted difference between
the real and generated feature means.

\subsection{Derivation of the Fisher Refinement Direction}
We derive the per-sample refinement direction induced by the Fisher objective and show its correspondence to the adversarial refinement gradient in \cref{eq:adv_gradient_decomposition}. During each generator update, the feature statistics used to construct the Fisher critic, including the class means and the Fisher direction $\mathbf{u}_{\ell}^{*}$, are treated as fixed. Thus, gradients propagate through the current generated feature $\mathbf{f}_{\ell}^{g}$ and the frozen teacher network, but not through the statistics defining the critic. For a single generated sample, the Fisher loss at layer $\ell$ is the logistic loss of the closed-form discriminator $D^{F}_{\ell}(\mathbf{f})=\sigma(r_{\ell}(\mathbf{f}))$,
\begin{equation}
    \ell_{\ell}^{F}
    =
    -\log D^{F}_{\ell}(\mathbf{f}_{\ell}^{g})
    =
    \log
    \left(
        1+e^{-r_{\ell}(\mathbf{f}_{\ell}^{g})}
    \right).
\end{equation}
Since
\begin{equation}
    \frac{d}{dr}\log\left(1+e^{-r}\right)
    =
    -\sigma(-r)
    =
    -\left(1-\sigma(r)\right),
\end{equation}
the gradient of the loss with respect to the generated sample is
\begin{equation}
    \nabla_{\mathbf{x}^{g}}\ell_{\ell}^{F}
    =
    -\left(
        1-D^{F}_{\ell}(\mathbf{f}_{\ell}^{g})
    \right)
    \nabla_{\mathbf{x}^{g}}
    r_{\ell}(\mathbf{f}_{\ell}^{g}).
\label{eq:supp_fisher_logistic_grad}
\end{equation}
Recall that the Fisher margin is
\begin{equation}
    r_{\ell}(\mathbf{f})
    =
    \mathbf{f}^{\top}\mathbf{u}_{\ell}^{*}
    -
    \frac{1}{2}
    \left(
        {\mathbf{u}_{\ell}^{*}}^{\top}\boldsymbol{\mu}_{\ell}^{r}
        +
        {\mathbf{u}_{\ell}^{*}}^{\top}\boldsymbol{\mu}_{\ell}^{g}
    \right).
\end{equation}
Because $\mathbf{u}_{\ell}^{*}$ and the class means are fixed during the
current generator update, the margin depends on $\mathbf{x}^{g}$ only through
$\mathbf{f}_{\ell}^{g}$. Therefore,
\begin{equation}
    \nabla_{\mathbf{f}_{\ell}^{g}}
    r_{\ell}(\mathbf{f}_{\ell}^{g})
    =
    \mathbf{u}_{\ell}^{*}.
\end{equation}
The generated feature $\mathbf{f}_{\ell}^{g}$ is obtained from the frozen
teacher using the noised generated sample
\begin{equation}
    \mathbf{x}_{t_F}^{g}
    =
    \alpha_{t_F}\mathbf{x}^{g}
    +
    \sigma_{t_F}\boldsymbol{\epsilon}.
\end{equation}
Applying the chain rule through the teacher feature gives
\begin{equation}
    \nabla_{\mathbf{x}_{t_F}^{g}}
    r_{\ell}(\mathbf{f}_{\ell}^{g})
    =
    \left(
        \frac{\partial \mathbf{f}_{\ell}^{g}}
             {\partial \mathbf{x}_{t_F}^{g}}
    \right)^{\!\top}
    \mathbf{u}_{\ell}^{*}.
\label{eq:supp_fisher_score_input_grad}
\end{equation}
Since
\begin{equation}
    \frac{\partial \mathbf{x}_{t_F}^{g}}
         {\partial \mathbf{x}^{g}}
    =
    \alpha_{t_F}\mathbf{I},
\end{equation}
we obtain
\begin{equation}
    \nabla_{\mathbf{x}^{g}}
    r_{\ell}(\mathbf{f}_{\ell}^{g})
    =
    \alpha_{t_F}
    \left(
        \frac{\partial \mathbf{f}_{\ell}^{g}}
             {\partial \mathbf{x}_{t_F}^{g}}
    \right)^{\!\top}
    \mathbf{u}_{\ell}^{*}.
\end{equation}
Substituting the margin gradient into
\cref{eq:supp_fisher_logistic_grad} gives the per-sample descent direction
\begin{equation}
    -\nabla_{\mathbf{x}^{g}}\ell_{\ell}^{F}
    =
    \underbrace{
        \left(
            1-D^{F}_{\ell}(\mathbf{f}_{\ell}^{g})
        \right)
    }_{\text{sample-dependent weight}}
    \underbrace{
        \alpha_{t_F}
    }_{\text{diffusion scaling}}
    \underbrace{
        \left(
            \frac{\partial \mathbf{f}_{\ell}^{g}}
                 {\partial \mathbf{x}_{t_F}^{g}}
        \right)^{\!\top}
        \mathbf{u}_{\ell}^{*}
    }_{\text{corrective direction}}.
\label{eq:supp_fisher_descent}
\end{equation}
This directly mirrors the adversarial refinement direction in
\cref{eq:adv_gradient_decomposition}. The factor
$1-D^{F}_{\ell}(\mathbf{f}_{\ell}^{g})=\sigma(-r_{\ell}(\mathbf{f}_{\ell}^{g}))$ is the
closed-form counterpart of the discriminator weight $1-D_{\psi}$.
The Fisher direction $\mathbf{u}_{\ell}^{*}$, propagated through the frozen
teacher, replaces the learned discriminator gradient, while
$\alpha_{t_F}$ provides the same diffusion scaling evaluated at the fixed
Fisher noise level. The gradient of the full Fisher objective is obtained by
taking the expectation of these per-sample gradients.

\begin{figure}
    \centering 
    \includegraphics[width=\columnwidth]{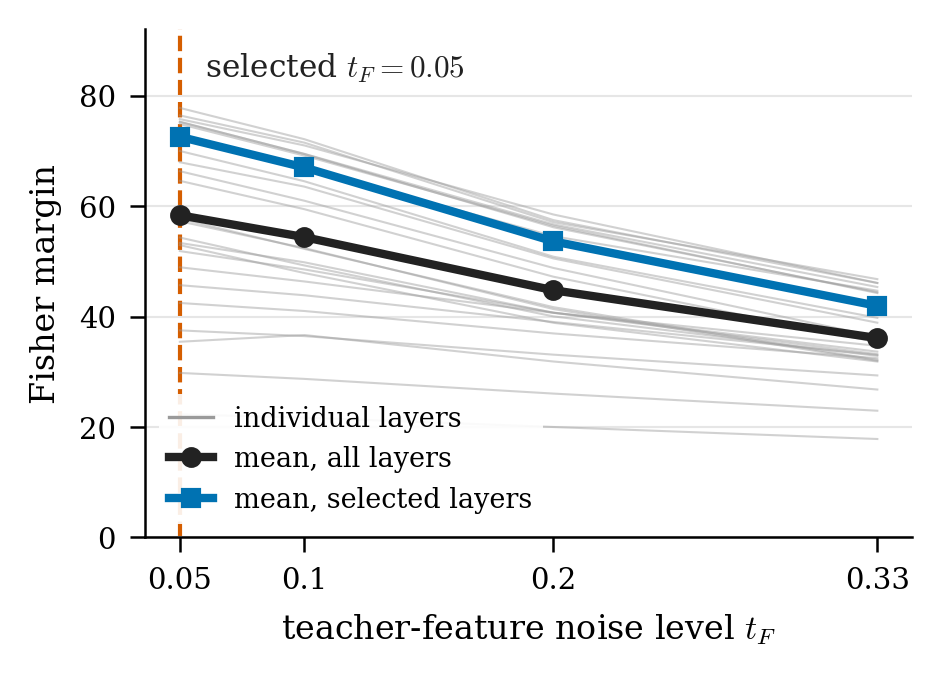}
    \caption{\textbf{Teacher Feature Noise Selection $t_F$ on SD3-Medium~\cite{sd3}.} Fisher separation margin between real images and Stage-I generations (5K each), computed on frozen SD3 teacher features at four noise levels. Grey: individual teacher layers. Black: mean over all layers. Blue: mean over the layers selected for FisherDistill. Separation decreases monotonically with $t_F$ for nearly all layers; we use $t_F = 0.05$.}
\label{fig:teacher_tf_selection_ablation}
\end{figure}

\section{Text-to-Image Evaluation Protocol}
For the DMD2~\cite{dmd2} setting, we follow its zero-shot COCO-2014~\cite{coco} protocol: $512 \times 512$ images generated with 30K validation captions, scored with clean-FID~\cite{clean_fid} against 40{,}504 validation images, with CLIP score~\cite{clipscore} computed using ViT-g-14~\cite{vit}. For 1.x-Distill~\cite{1_x_distill}, we follow its SD3-Medium~\cite{sd3} protocol: 10K COCO-2014~\cite{coco} prompts at $1024 \times 1024$. 1.x-Distill~\cite{1_x_distill} computes FID~\cite{clean_fid,fid} against
teacher-generated images. In contrast, we compute FID against real images, which provides a direct measure of distributional fidelity to the target data distribution and follows the standard evaluation protocol used in prior generative modeling literature~\cite{dmd, dmd2, ldm}. For CLIP score, we use ViT-B/32~\cite{vit}. For SiD-DiT~\cite{SiD-Dit}, we follow its 10K-COCO~\cite{coco} protocol at the SANA~\cite{sana} native resolution. Human-preference metrics (Aesthetics~\cite{aesthetics_score}, HPSv2~\cite{hpsv2}, ImageReward~\cite{imgrwrd}, PickScore~\cite{pick-a-pic}) are computed on 2{,}048 Pick-a-Pic~\cite{pick-a-pic} validation prompts for DMD2 and COCO-10K~\cite{coco} for the rest.

\section{Additional Ablations}
\paragraph{Selecting the teacher-feature noise level.} We choose \(t_F\) by measuring real-generated separability in frozen teacher features across candidate noise levels and selecting a low-noise regime that preserves strong separation across the layers used by FisherDistill. Shown here for SD3~\cite{sd3} (\cref{fig:teacher_tf_selection_ablation}), we measure the Fisher separation margin between real images and Stage-I generations using 5K samples per set, \(M_{\ell}(t_F)=\left|{\mathbf{u}_{\ell}^{*}}^{\top}(\boldsymbol{\mu}_{\ell}^{g}-\boldsymbol{\mu}_{\ell}^{r})\right|/\sqrt{{\mathbf{u}_{\ell}^{*}}^{\top}\boldsymbol{\Sigma}_{\ell}\mathbf{u}_{\ell}^{*}}\). Gray denotes individual teacher layers, black the mean across all layers, and blue the mean across the layers selected for FisherDistill. Separation decreases with increasing \(t_F\) for nearly all layers as noise suppresses residual real-generated differences. We therefore use \(t_F=0.05\) for SD3~\cite{sd3}. We similarly decide the noise level of 0.085 for EDM~\cite{edm}, 0.2 for the SANA~\cite{sana} models and $t=21$ for SD1.5~\cite{ldm}.

\section{ImageNet Results}
\begin{figure*}[t]
    \centering 
    \includegraphics[width=\textwidth]{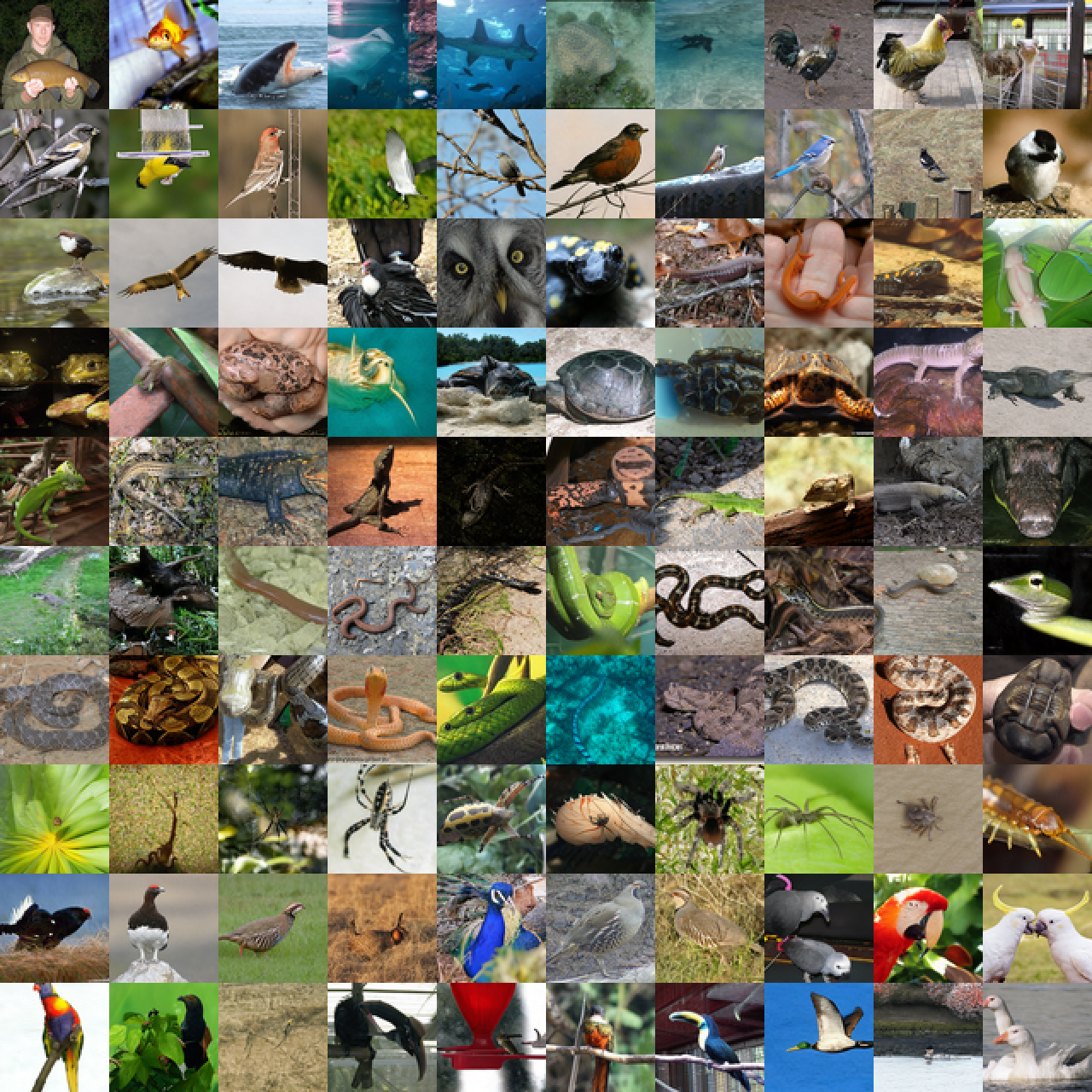}
    \caption{\textbf{Uncurated one-step samples} from GAN refined~\cite{dmd2} generator on ImageNet
$64$$\times$$64$ (FID 1.42) distilled from EDM~\cite{edm} in DMD~\cite{dmd}, one sample per class for classes 0-99.}
    \label{fig:imagenet_suppl_dmd_gan}
\end{figure*}

\begin{figure*}[t]
    \centering 
    \includegraphics[width=\textwidth]{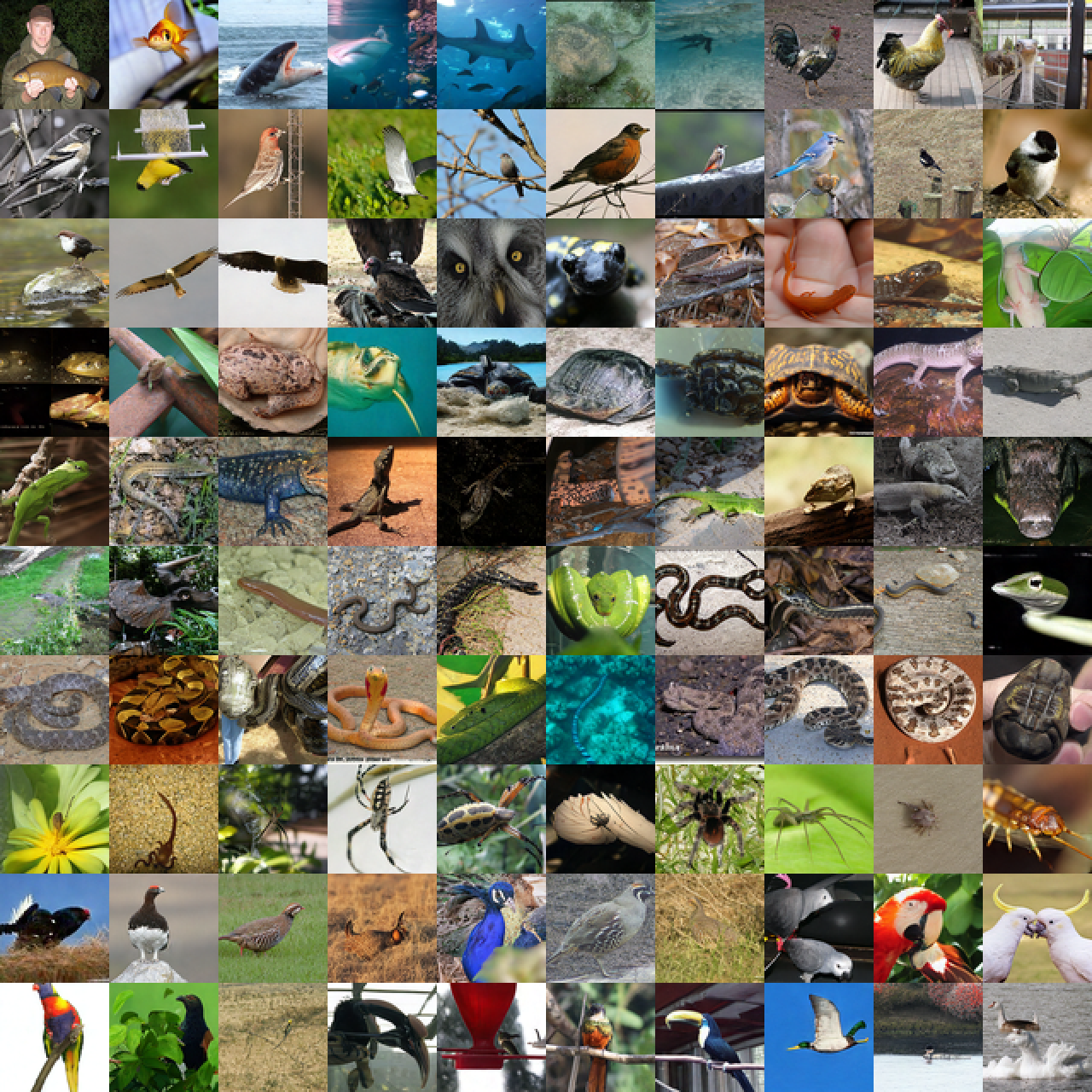}
    \caption{\textbf{Uncurated one-step samples} from FisherDistill generator on ImageNet
$64$$\times$$64$ (FID 1.40) distilled from EDM~\cite{edm} in DMD~\cite{dmd}, one sample per class for classes 0-99.}
    \label{fig:imagenet_suppl_dmd_ours}
\end{figure*}

\begin{figure*}[t]
    \centering 
    \includegraphics[width=\textwidth]{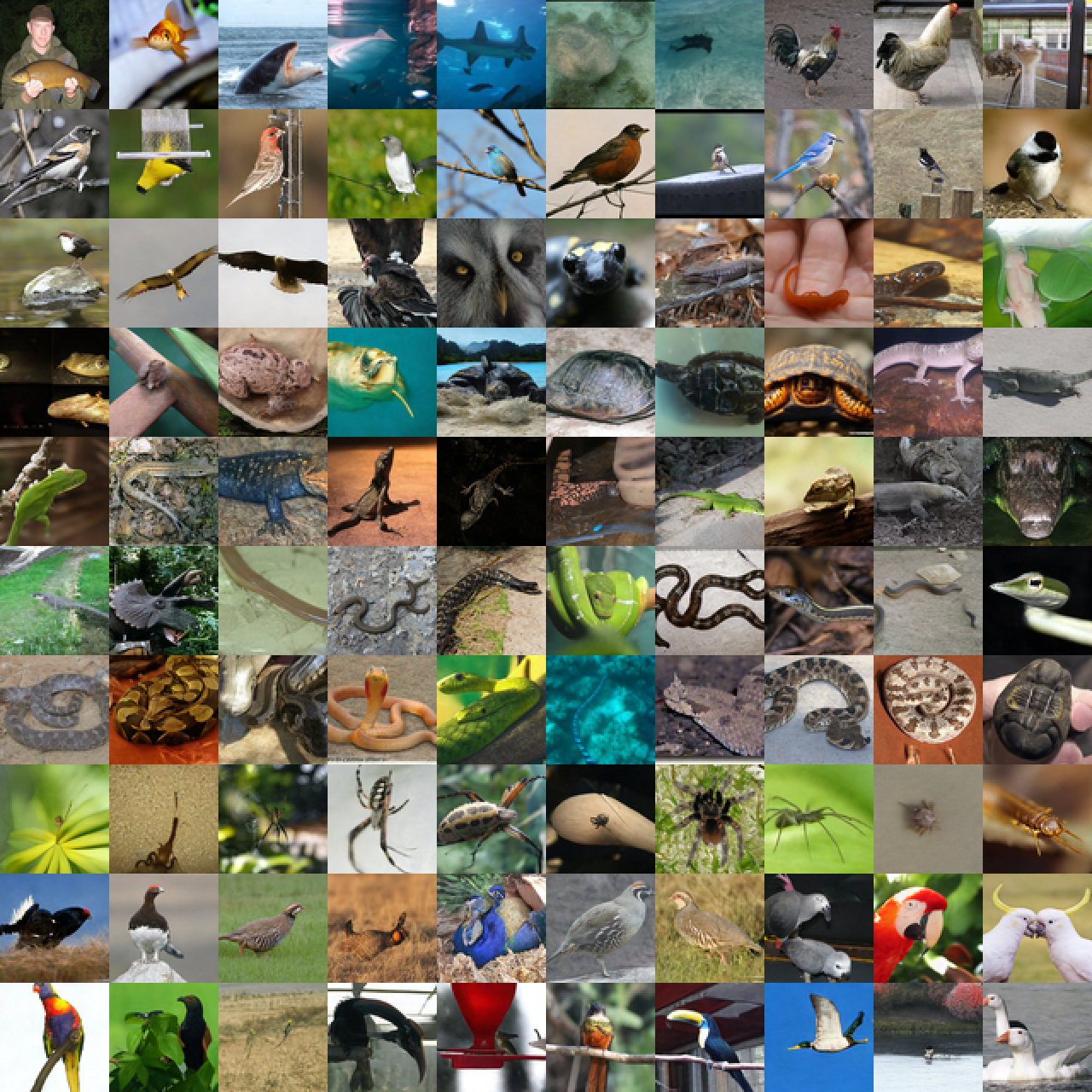}
    \caption{\textbf{Uncurated one-step samples} from RealUID~\cite{realuid} generator on ImageNet
$64$$\times$$64$ (FID 3.07) distilled from EDM~\cite{edm} in DMD~\cite{dmd}, one sample per class for classes 0-99.}
    \label{fig:imagenet_suppl_dmd_realuid}
\end{figure*}

Results on 100 ImageNet classes in their original $64\times64$ dimensions distilled from EDM are in~\cref{fig:imagenet_suppl_dmd_gan,fig:imagenet_suppl_dmd_ours,fig:imagenet_suppl_dmd_realuid} for GAN refined~\cite{dmd2}, FisherDistill refined and RealUID~\cite{realuid} refined generators.

\end{document}